\documentclass[conference]{IEEEtran}
\IEEEoverridecommandlockouts
\usepackage{cite}
\usepackage{amsmath,amssymb,amsfonts}
\usepackage{algorithmic}
\usepackage{graphicx}
\usepackage{textcomp}
\usepackage{xcolor}
\usepackage{booktabs}
\usepackage{graphicx}
\usepackage{subcaption}
\usepackage{adjustbox}
\usepackage[table]{xcolor}
\usepackage{multirow}
\usepackage{adjustbox}
\usepackage{bbm}
\usepackage[ruled,vlined]{algorithm2e}
\usepackage{hyperref}  
\usepackage[most]{tcolorbox}
\def\BibTeX{{\rm B\kern-.05em{\sc i\kern-.025em b}\kern-.08em
    T\kern-.1667em\lower.7ex\hbox{E}\kern-.125emX}}
\begin{document}

\title{Cross-Domain Generalization of Multimodal LLMs for Global Photovoltaic Assessment\\
}

\author{\IEEEauthorblockN{Muhao Guo}
\IEEEauthorblockA{\textit{School of Electrical, Computer and Energy Engineering} \\
\textit{Arizona State University}\\
Tempe, United States \\
mguo26@asu.edu}
\and
\IEEEauthorblockN{Yang Weng}
\IEEEauthorblockA{\textit{School of Electrical, Computer and Energy Engineering} \\
\textit{Arizona State University}\\
Tempe, United States \\
Yang.Weng@asu.edu}
}


\maketitle
\begin{abstract}
The rapid expansion of distributed photovoltaic (PV) systems poses challenges for power grid management, as many installations remain undocumented. While satellite imagery provides global coverage, traditional computer vision (CV) models such as CNNs and U-Nets require extensive labeled data and fail to generalize across regions. This study investigates the cross-domain generalization of a multimodal large language model (LLM) for global PV assessment. By leveraging structured prompts and fine-tuning, the model integrates detection, localization, and quantification within a unified schema. Cross-regional evaluation using the $\Delta$F1 metric demonstrates that the proposed model achieves the smallest performance degradation across unseen regions, outperforming conventional CV and transformer baselines. These results highlight the robustness of multimodal LLMs under domain shift and their potential for scalable, transferable, and interpretable global PV mapping.
\end{abstract}


\section{Introduction}
The rapid deployment of distributed photovoltaic (PV) systems has transformed the landscape of modern power distribution networks \cite{guo2023graph}. Accurate visibility of small-scale and residential PV installations is essential for power grid operators to perform reliable load forecasting \cite{li2025external, li2025exarnn, li2025latent}, hosting capacity analysis\cite{wu2022spatial,yuan2022determining}, and distributed energy resource (DER) management\cite{chen2023optimal}. However, many of these installations remain undocumented in official records. Existing databases, such as those maintained by the U.S. Energy Information Administration (EIA) and state interconnection registries, are typically updated infrequently and lack sufficient spatial granularity. This information gap limits situational awareness in planning and operation of active distribution networks and microgrids.

Satellite imagery provides a globally consistent and continuously updated source of information for PV assessment. Yet traditional computer vision (CV) approaches, including convolutional neural networks (CNNs) \cite{simonyan2014very} and U-Net-based segmentation models \cite{bouazizintegrated}, encounter several critical challenges. They demand large volumes of labeled data \cite{luo2022solar}, struggle to distinguish visually similar surfaces such as dark roofs and parking lots, and exhibit poor generalization when transferred to regions with different rooftop geometries or illumination conditions. Furthermore, these models are designed for narrow, single-task objectives and lack interpretability, making their predictions difficult to validate and integrate into decision-making processes for power systems.

Recent advances in LLMs have opened a new direction for multimodal learning \cite{guo2023msq}. By jointly reasoning over text and imagery, LLMs are capable of performing high-level semantic interpretation, contextual reasoning, and adaptive decision-making with minimal supervision. Such capabilities make them particularly promising for scalable solar infrastructure mapping. In earlier work, the PV Assessment with LLMs (PVAL) \cite{guo2025solar} framework demonstrated that fine-tuned multimodal LLMs can detect, localize, and quantify rooftop PV installations with competitive accuracy using structured prompts and auto-labeled datasets. While this line of research established the feasibility of LLM-based solar detection, questions remain regarding their robustness and adaptability across geographically and climatically diverse environments.

\begin{figure}[t]
\centering
\includegraphics[width=1\columnwidth]{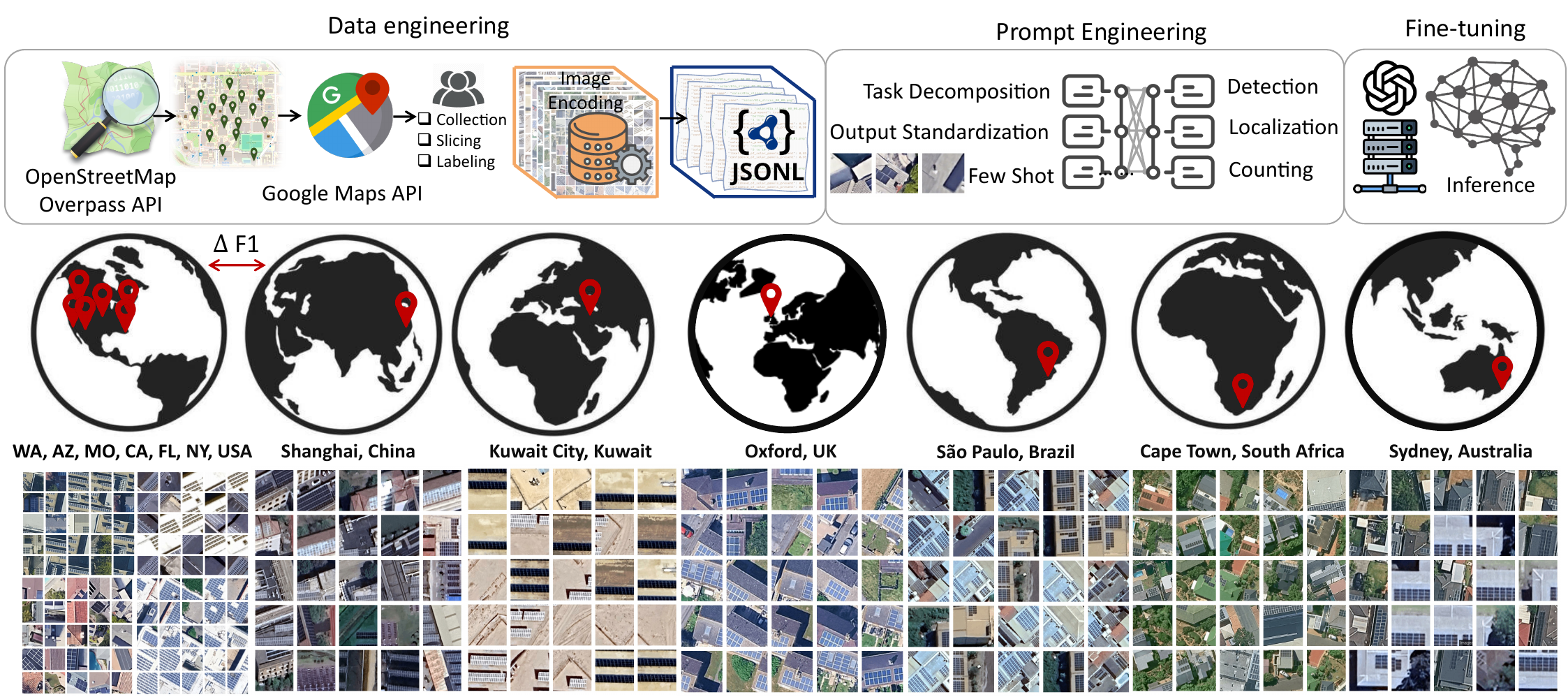}
\caption{Framework of the proposed multimodal LLM for global photovoltaic (PV) assessment. The pipeline includes data engineering, prompt engineering, and fine-tuning stages. Rooftop imagery from seven global regions is used to evaluate cross-domain generalization, with $\Delta$F1 representing performance changes between the fine-tuning and unseen regions.}
\vspace{-1em}
\label{fig:Framework}
\end{figure}

This paper investigates the problem of domain adaptation for multimodal LLMs in global solar infrastructure mapping. Specifically, we examine how a model fine-tuned on data from U.S. regions performs when applied to unseen domains worldwide, and we analyze how performance degrades with increasing domain shift (See Fig. \ref{fig:Framework}). 
Through experiments conducted on cities across six continents, we quantify accuracy, calibration, and cross-domain transferability. The results demonstrate that multimodal LLMs maintain high reliability even under significant visual and climatic variations, highlighting their potential as a foundation for globally scalable, interpretable, and data-efficient PV monitoring.

\section{Problem Formulation}
Let a satellite image be denoted as \( I \). The objective is to construct a mapping function 
\begin{equation}
    f(I) \rightarrow \{ D, L, Q\},
\end{equation}
where \( D \) indicates the binary detection of PV presence, \( L \) specifies spatial localization within the image, and \( Q \) estimates the panel count or surface area. 
The function \( f(\cdot) \) should generalize across diverse geographical domains and architectural contexts without exhaustive retraining.

In this work, we formulate PV detection as a multimodal inference problem leveraging LLMs conditioned on both imagery and structured textual prompts. The proposed framework employs fine-tuning and few-shot learning re-anchoring to adapt a base LLM for PV infrastructure mapping. To quantitatively assess the model’s robustness to domain shift, we introduce the $\Delta$\textit{F1} metric, which measures generalization performance across geographical regions. 
\begin{equation}
\Delta \textit{F1} = \textit{F1}_{\textit{target region}} - \textit{F1}_{\textit{training region}},
\end{equation}
where \(\textit{F1}_{\textit{target region}}\) and \(\textit{F1}_{\textit{training region}}\) denote F1-Scores on unseen and source domains, respectively. A positive $\Delta$F1 indicates improved transferability, while a negative value reflects performance degradation under domain shift. This metric provides a standardized means to evaluate how well the model generalizes beyond its fine-tuning region.

\section{Methodology}
This section outlines the proposed framework for cross-domain generalization of multimodal LLMs in global PV mapping. Building on the PVAL foundation, the framework is designed to evaluate model robustness under diverse spatial, climatic, and architectural conditions. It consists of three main components: data engineering, multimodal LLM configuration, and domain-shift evaluation.

\subsection{Data Engineering}
Data engineering forms the foundation of PVAL, ensuring that the model is trained and evaluated on high-quality, well-structured, and reproducible datasets. The data pipeline comprises three main components: data collection, image slicing, and annotation.
Geographic coordinates of PV installations were obtained from OpenStreetMap (OSM) using the Overpass API\footnote{\url{https://overpass-api.de/}}. Each region’s query, endpoint, and export timestamp were recorded to preserve the exact OSM snapshot. These coordinates anchored the retrieval of satellite imagery from the Google Maps Static API\footnote{\url{https://developers.google.com/maps/documentation/maps-static/overview}}, with fixed parameters (\texttt{maptype=satellite}, \texttt{zoom=20}, and \texttt{size=400x400}). 
To capture a wide range of environmental and architectural conditions, the dataset covers six U.S. regions (Seattle, WA; Orlando, FL; Osage Beach, MO; Harlem, NY; Tempe, AZ; and Santa Ana, CA) and six international cities (Sydney, Australia; Cape Town, South Africa; Kuwait City, Middle East; Oxford, UK; São Paulo, Brazil; and Shanghai, China), supporting cross-regional and cross-continental generalization.

\begin{figure}[htbp]
\centering
\includegraphics[width=0.85\columnwidth]{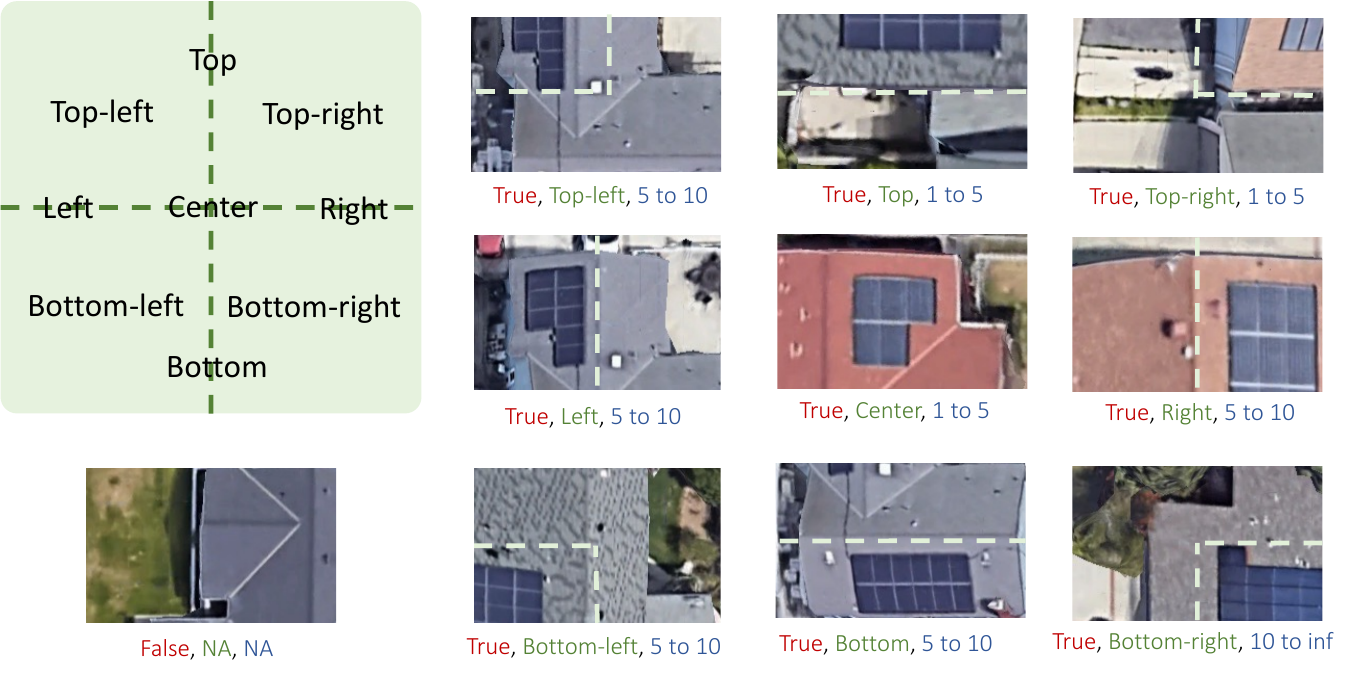}
\caption{Solar panel labeling schema and representative examples. The left panel shows nine spatial regions (top, bottom, left, right, center, and four diagonals), plus the “NA” case indicating no solar panels. 
The right panel displays annotated rooftop tiles with labels specifying (i) presence (True/False), (ii) location, and (iii) quantity range. 
}
\vspace{-0.5em}
\label{fig:examples}
\end{figure}

Each high-resolution image was divided into a $4\times4$ grid, yielding 16 tiles per image. This deterministic tiling strategy expanded the dataset while maintaining spatial resolution and ensuring visibility of small-scale PV arrays. It enabled the model to jointly learn presence detection (D), localization (L), and quantity estimation (Q). Each tile was manually annotated by trained PV specialists for solar panel presence, location, and quantity. Fig. ~\ref{fig:examples} illustrates the standardized labeling schema for the three tasks, along with representative annotated examples. This human-in-the-loop process provides high-quality ground truth for model fine-tuning and evaluation \cite{blasch2012high}.

The final dataset integrates both U.S. and global imagery from open sources, ensuring transparency and reproducibility. Each image tile is labeled for \emph{solar panel presence}, with additional structured annotations for \emph{location} and \emph{quantity} (intervals: $(0,1]$, $(1,5]$, $(5,10]$, $(10,+\infty)$, or ``NA"). These labels provide a standardized and interpretable representation of rooftop PV characteristics across diverse regions.

\begin{table}[h]
\centering
\caption{Summary of training and evaluation datasets.
}
\renewcommand{\arraystretch}{1.1}
\setlength{\tabcolsep}{5pt}
\begin{adjustbox}{width=\columnwidth,center}
\begin{tabular}{llcc}
\hline
\textbf{Continent} & \textbf{Region / City} & \textbf{Role} & \textbf{Size} \\
\hline
\multirow{6}{*}{\textbf{North America}} 
 & Santa Ana, CA & Training (fine-tuning) & 2{,}000 tiles \\
 & Tempe, AZ + Santa Ana, CA & Large-scale Test & $\sim$100{,}000 tiles \\ 
 & Seattle, WA & Cross-regional Test & 480 tiles \\ 
 & Orlando, FL & Cross-regional Test & 480 tiles \\ 
 & Osage Beach, MO & Cross-regional Test & 480 tiles \\ 
 & Harlem, NY & Cross-regional Test & 480 tiles \\ 
\hline
\textbf{Oceania} & Sydney, Australia & Cross-continental Test & 480 tiles \\ 
\textbf{Africa} & Cape Town, South Africa & Cross-continental Test & 480 tiles \\ 
\textbf{Asia (Middle East)} & Kuwait City, Kuwait & Cross-continental Test & 480 tiles \\ 
\textbf{Asia (East Asia)} & Shanghai, China & Cross-continental Test & 480 tiles \\ 
\textbf{Europe} & Oxford, UK & Cross-continental Test & 480 tiles \\ 
\textbf{South America} & S\~{a}o Paulo, Brazil & Cross-continental Test & 480 tiles \\ 
\hline
\end{tabular}
\end{adjustbox}
\label{tab:dataset_summary}
\end{table}

Table~\ref{tab:dataset_summary} summarizes the datasets used for training and evaluation. Both baseline models and the PVAL framework were fine-tuned on 2{,}000 annotated tiles from Santa Ana, CA. The large-scale evaluation set includes about 100{,}000 tiles from Tempe and Santa Ana, while 480 tiles per region were used for cross-domain generalization tests across diverse climates and geographies.

\subsection{Multimodal LLM Configuration}
Configuring the PVAL system for solar panel detection involves a multi-faceted approach that integrates prompt engineering, output standardization, and supervised fine-tuning. This configuration is critical for steering the foundational GPT-4o model towards the specific, high-precision task of geospatial analysis.

\label{sec:llm_config}
\begin{tcolorbox}[
    colback=white,
    colframe=gray!75!black,
    title=\tiny\bfseries Prompt,
    fonttitle=\bfseries,
    width=1\linewidth,
    boxrule=0.5mm,
    left=1.5mm,
    right=0mm,
    top=0mm,
    bottom=-2mm
]
\tiny\ttfamily
\label{prompt:structure}
\colorbox{blue!10}{\textbf{Task Decomposition}} \\
Identify the presence of solar panels in images of residential rooftops, and determine their locations and quantity within the images. 
You will be provided with images that may contain residential rooftop solar systems. Analyze each image to detect solar panels.
\\
\textbf{Steps:} \\
1. **Image Analysis**: Examine the entire image to identify any objects that appear to be solar panels.\\
2. **Panel Location**: Determine the coordinates or area within the image where the solar panels are located.\\
3. **Panel Quantification**: Calculate or estimate the number of solar panels based on their appearance and arrangement.\\
\colorbox{yellow!10}{\textbf{Output Standardization}} \\
The output should be in JSON format, structured as follows, with each field restricted to specific possible values for consistency and accuracy: \\
"solar\_panels\_present": A boolean value indicating if solar panels are detected. \\
Possible values: [true, false] \\
"location": A description or coordinates indicating where the panels are located within the image. \\
Possible values: [left, right, bottom, top, top-left, top-right, bottom-right, bottom-left, center, NA] \\
"quantity": The number of solar panels detected in the image. \\
Possible values: [0 to 1, 1 to 5, 5 to 10, 10 to inf, NA] \\
"likelihood\_of\_solar\_panels\_present": A value indicating the probability of solar panels being present. \\
Possible values: A decimal range from 0.00 to 1.00 \\
"confidence\_of\_solar\_panels\_present": A value indicating the model's confidence in its prediction. \\
Possible values: A decimal range from 0.00 to 1.00 \\
\colorbox{orange!10}{\textbf{Few-shot Prompting}} \\
$\cdot$ Example 1 (Solar): \\
\{ "solar\_panels\_present": true, \\
\mbox{} \mbox{} "location": "top-left", \\
\mbox{} \mbox{} "quantity": "0 to 1", \\
\mbox{} \mbox{} "likelihood\_of\_solar\_panels\_present": 0.98, \\
\mbox{} \mbox{} "confidence\_of\_solar\_panels\_present": 0.90 \} \\
$\cdot$ Example 2 (No Solar): \\
\{ "solar\_panels\_present": false, \\
\mbox{} \mbox{} "location": "NA", \\
\mbox{} \mbox{} "quantity": "NA", \\
\mbox{} \mbox{} "likelihood\_of\_solar\_panels\_present": 0.21, \\
\mbox{} \mbox{} "confidence\_of\_solar\_panels\_present": 0.87 \} \\
\end{tcolorbox}

The process begins with prompt engineering, which defines the model's operational contract. As detailed in the system prompt (see Prompt~\ref{prompt:structure}), this is a comprehensive guide. It provides ``task decomposition" (analysis, location, quantification) and, crucially, enforces ``output standardization" by demanding a strict JSON schema with predefined value ranges (e.g., ``top-left", ``10 to inf"). This schema-guided approach is reinforced with ``few-shot learning" examples, ensuring the model's responses are consistent and machine-readable.

This prompt structure is then used as the basis for supervised fine-tuning, the procedure for which is outlined in Algorithm~\ref{algorithm:finetune_simple}. The training dataset is prepared in JSON Lines (JSONL) format, a format well-suited for the OpenAI API. In this dataset, each entry pairs a satellite image (encoded in \textit{base64} \cite{josefsson2006base16}) with its corresponding ground truth JSON output, which adheres to the exact schema defined in the prompt. 

During the fine-tuning job, we use OpenAI's supervised fine-tuning API with image-conditioned prompts and JSON-formatted targets. The training objective is the standard autoregressive language-model cross-entropy:
\begin{equation}
\mathcal{L}_{\text{LM}} = -\frac{1}{N}\sum_{i=1}^{N}\frac{1}{T_i}\sum_{t=1}^{T_i}
\log p_\theta\!\left(y_{i,t}\mid y_{i,<t},\,x_i\right),
\label{eq:lm_loss}
\end{equation}
where $x_i$ denotes the input (image + text), $y_{i,1:T_i}$ are the target tokens of the JSON output, and $p_\theta$ is the model distribution.
This setup leverages GPT-4o’s multimodal reasoning while aligning it to the photovoltaic assessment task under a schema-constrained JSON output.

\begin{algorithm}[t]
\caption{Fine-Tuning Procedure}
\label{algorithm:finetune_simple}
\DontPrintSemicolon
\SetAlCapFnt{\small}\SetAlCapNameFnt{\small}
\scriptsize
\KwIn{Training file \text{data.jsonl}, base model \text{GPT-4o}, API key, hyperparameters: epochs, batch size, learning rate (lr), temperature (T) }
\KwOut{Fine-tuned model $\theta^\ast$}
\textbf{1) Initialize client:} $\text{client} \leftarrow \text{OpenAI(api\_key)}$\\
\textbf{2) Upload training file:} 
$\text{file\_id} \leftarrow \text{client.files.create(file=``data.jsonl", purpose=``fine-tune").id}$\\
\textbf{(3) Create fine-tuning job:} 
$\text{job} \leftarrow \text{client.fine\_tuning.jobs.create(}$
\hspace{1em} 
$\text{training\_file=file\_id, model=base\_model, n\_epochs, batch, lr)}$
\\
\textbf{4) Monitor job status:} \While{$\text{status} \notin \{\text{succeeded}, \text{failed}\}$}{
  $\text{job} \leftarrow \text{client.fine\_tuning.jobs.retrieve(job.id)}$;\\
  $\text{status} \leftarrow \text{job.status}$;
}
\textbf{5) On success:} $\theta^\ast \leftarrow \text{job.fine\_tuned\_model}$\\
\textbf{6) Use fine-tuned model:}\\
$\text{response} \leftarrow \text{client.chat.completions.create(} \text{model=}\theta^\ast, \text{messages=[\{user prompt\}]}, T{=}0)$\\
\Return $\theta^\ast$
\end{algorithm}

\subsection{Evaluation Metrics}
\label{sec:Evaluation_Metrics}
The model outputs schema-constrained categorical values rather than pixel masks, so IoU and mAP are inapplicable. We report schema-aligned metrics. For the binary solar-panel presence task, with labels $y_i \in \{0,1\}$ and predictions $\hat{y}_i \in \{0,1\}$, let $TP$, $FP$, and $FN$ denote true positives, false positives, and false negatives, respectively:
$
\mathrm{Precision}= {TP}/(TP+FP), \ 
\mathrm{Recall}={TP}/(TP+FN), \
F1= 2\cdot \mathrm{Precision}\cdot\mathrm{Recall} / (\mathrm{Precision}+\mathrm{Recall}).
$
For categorical attributes such as location and quantity, performance is measured by exact-match accuracy,
$\mathrm{Accuracy}=\frac{1}{N}\sum_{i=1}^{N}\mathbbm{1}\!\left[\hat{y}_i=y_i\right].$
Cross-domain performance is summarized by the difference in F1 between the source and target domains:
$\Delta \textit{F1} = \textit{F1}_{\textit{target}} - \textit{F1}_{\textit{source}}$.

\section{Numerical Results}
\subsection{Experiment Setup}
Fine-tuning and inference were conducted using the OpenAI GPT-4o multimodal architecture through its API interface. The training dataset was prepared in JSONL format, containing paired satellite imagery and structured text prompts, as described in Section~III. Fine-tuning was performed for five epochs on OpenAI’s managed infrastructure to adapt the base model for PV detection, localization, and quantification tasks. 

For benchmarking and large-scale inference, all experiments were executed on the \textit{Sol} supercomputing cluster at Arizona State University. Each node is equipped with NVIDIA~A100 GPUs (80~GB memory, CUDA~12.7), AMD~EPYC~7413 CPUs, and 256~GB of system RAM. Baseline computer vision models, including U-Net \cite{bouazizintegrated}, ResNet-152\cite{he2016deep}, Inception-v3\cite{szegedy2016rethinking}, VGG-19\cite{simonyan2014very}, and ViT-Base-16\cite{dosovitskiy2020image}, were implemented in Python~3.9 with identical preprocessing, data augmentation, and batch configurations to ensure a fair comparison with the proposed LLM-based framework.
For reference, the baseline convolutional and transformer-based models were trained for ten epochs on the same source-domain dataset using binary cross-entropy loss and the Adam optimizer (learning rate $\eta = 10^{-3}$, batch size = 16). The proposed multimodal LLM framework was evaluated on the same dataset splits to ensure consistency and comparability across all experiments.

\subsection{Comparative Performance Analysis}

The initial phase of our evaluation focused on establishing the baseline performance of the PVAL against a suite of traditional CNN-based and transformer-based models. Take Santa Ana as an example, the fine-tuned PVAL demonstrates superior performance across all standard classification metrics. 
\begin{figure}[htbp]
\centering
\includegraphics[width=0.9\columnwidth]{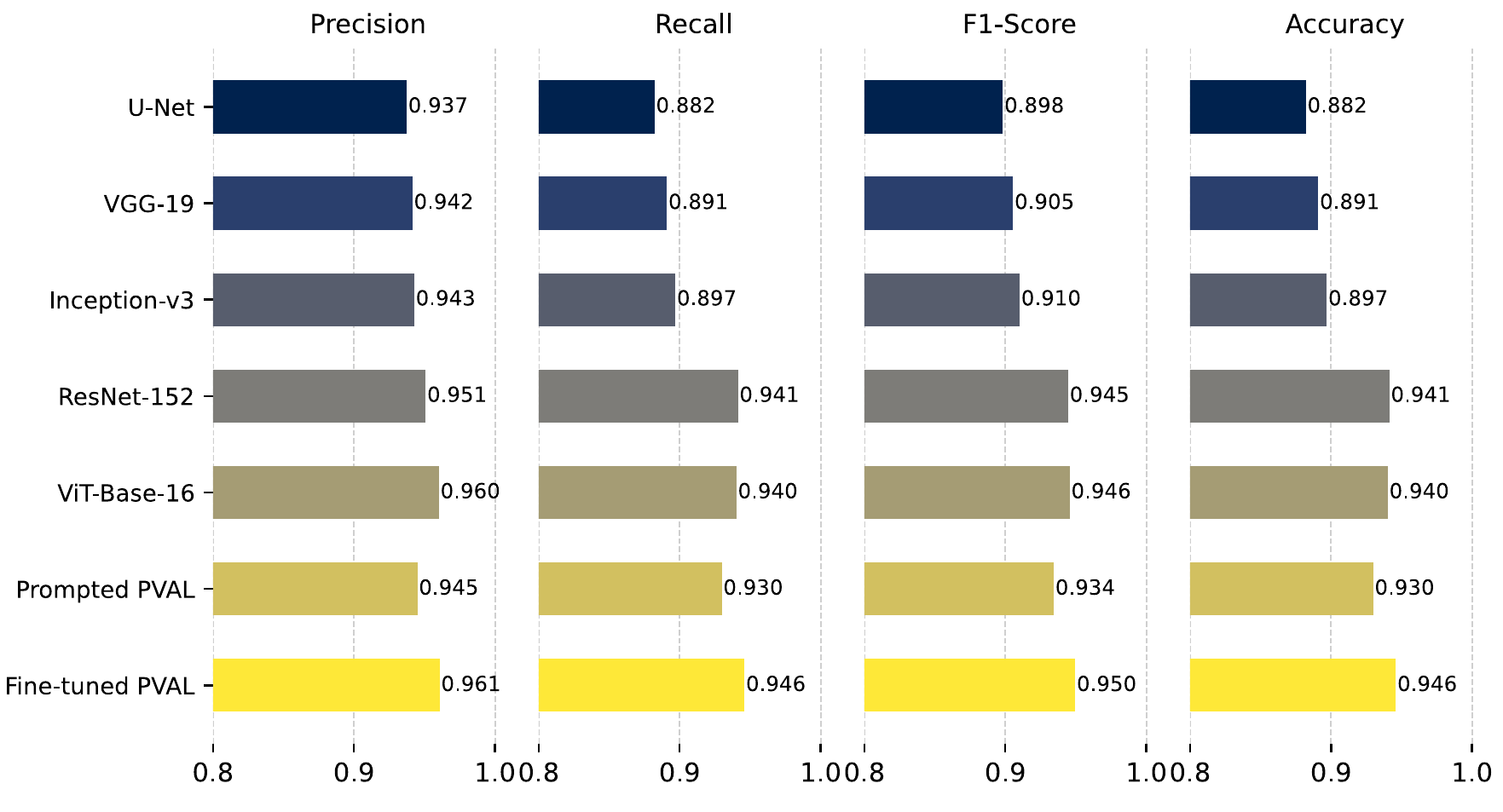}
\caption{Comparison of model performance across Precision, Recall, F1-Score, and Accuracy metrics. 
}
\vspace{-0.5em}
\label{fig:performance_baselines}
\end{figure}
As illustrated in Fig.~\ref{fig:performance_baselines}, it achieves the highest Precision (0.961), Recall (0.946), F1-Score (0.950), and Accuracy (0.946), outperforming established architectures like U-Net, ResNet-152, and ViT-Base-16. This superiority, especially in precision, stems from the LLM's underlying architecture. Unlike pure CV models that learn pixel-level patterns from scratch, the multimodal LLM leverages vast pre-training on world knowledge. It possesses a high-level semantic understanding of constituent concepts (e.g., ``roof," ``shadow," ``skylight," ``solar panel"). This allows it to more effectively disambiguate visually similar but semantically different objects, such as mistaking a dark skylight for a solar panel, a common failure case for traditional models. The ``Prompted PVAL" baseline's strong showing further underscores this, though fine-tuning is clearly critical for specializing this knowledge.

\begin{figure}[htbp]
\centering
\includegraphics[width=0.75\columnwidth]{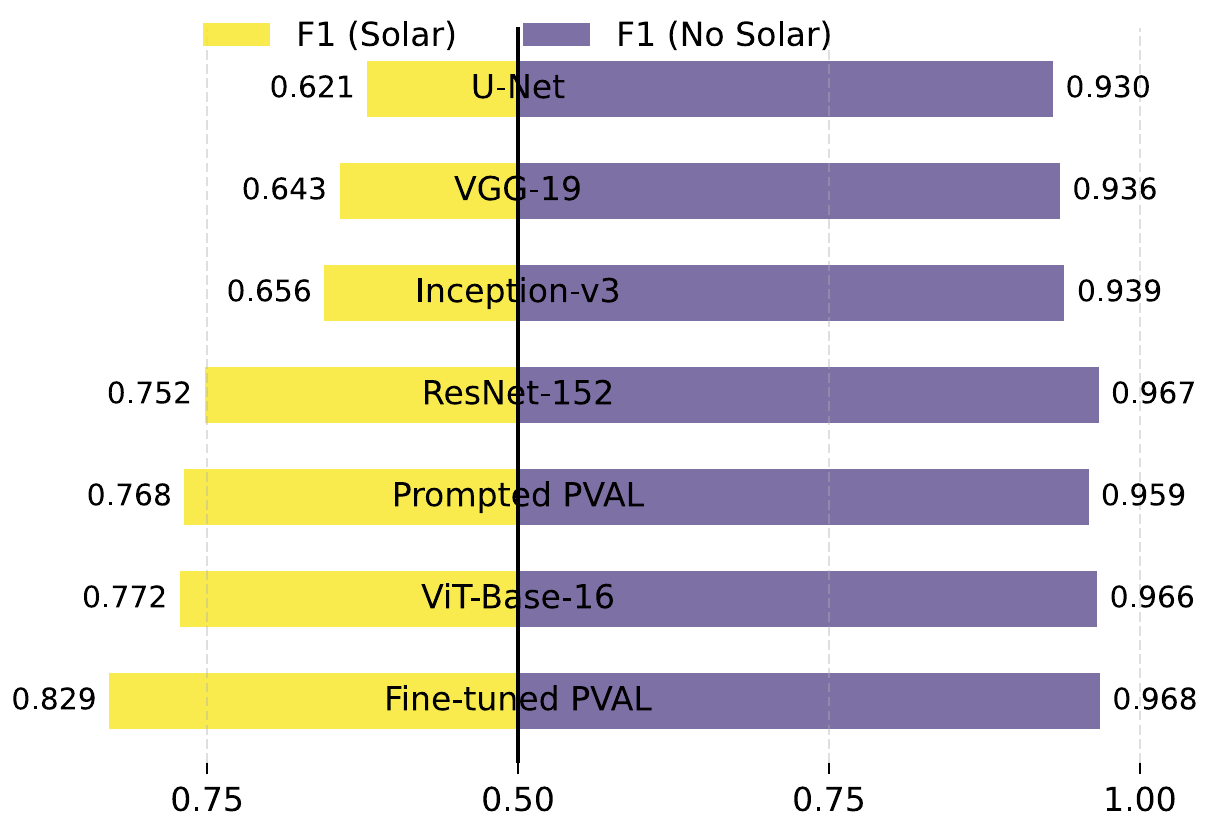}
\caption{Comparison of F1-scores for solar and non-solar datasets across different models. 
}
\vspace{-1.5em}
\label{fig:performance_solar_nosolar}
\end{figure}

We further analyzed the model's robustness to class imbalance in Fig.~\ref{fig:performance_solar_nosolar}, which separates the F1-Score for ``Solar" (positive class) and ``No Solar" (negative class) images. While all models demonstrate high proficiency in identifying the common ``No Solar" tiles, the primary challenge lies in correctly identifying the ``Solar" instances. Here, the PVAL achieves an F1-Score of 0.829 for the ``Solar" class, decisively leading the next-best model (ViT-Base-16 at 0.772). This suggests that traditional CV models, when trained on an imbalanced dataset, tend to overfit to the majority class, treating the minority ``Solar" class as a statistical anomaly. The LLM, guided by its structured prompt and robust feature representation, maintains a more balanced and robust predictive capability, minimizing false negatives without sacrificing precision.

\begin{figure}[htbp]
\centering
\includegraphics[width=0.8\columnwidth]{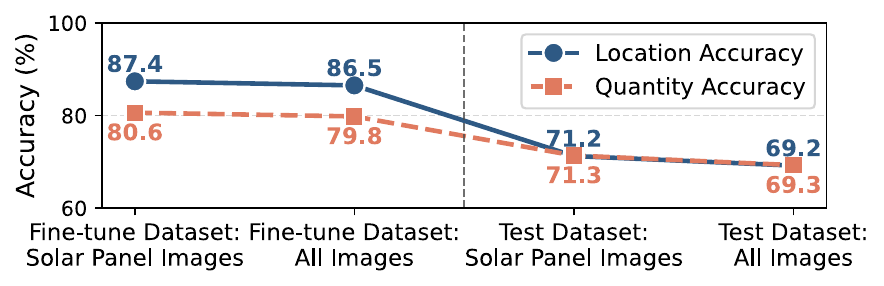}
\caption{
Accuracy comparison of solar panel location and quantity detection for fine-tuning and test datasets. 
}
\vspace{-0.5em}
\label{fig:Location_Count}
\end{figure}

Beyond binary classification, we evaluated the PVAL framework's advanced multi-task capabilities in performing location and quantity estimation, with results presented in Fig.~\ref{fig:Location_Count}.
On the fine-tune dataset, the model achieved high accuracy for both Location (87.4\%) and Quantity (80.6\%). This task is a key differentiator for the LLM. For a CV model like U-Net, deriving ``location" and ``quantity" would require brittle and complex post-processing logic (e.g., contour detection, object counting) on a segmented pixel mask. The PVAL generates this information natively as structured JSON output. This demonstrates a capacity for genuine reasoning: quantifying and spatializing, rather than simple pixel-wise segmentation.

\subsection{Cross-Domain Performance Analysis}

\begin{figure}[htbp]
\centering
\includegraphics[width=1\columnwidth]{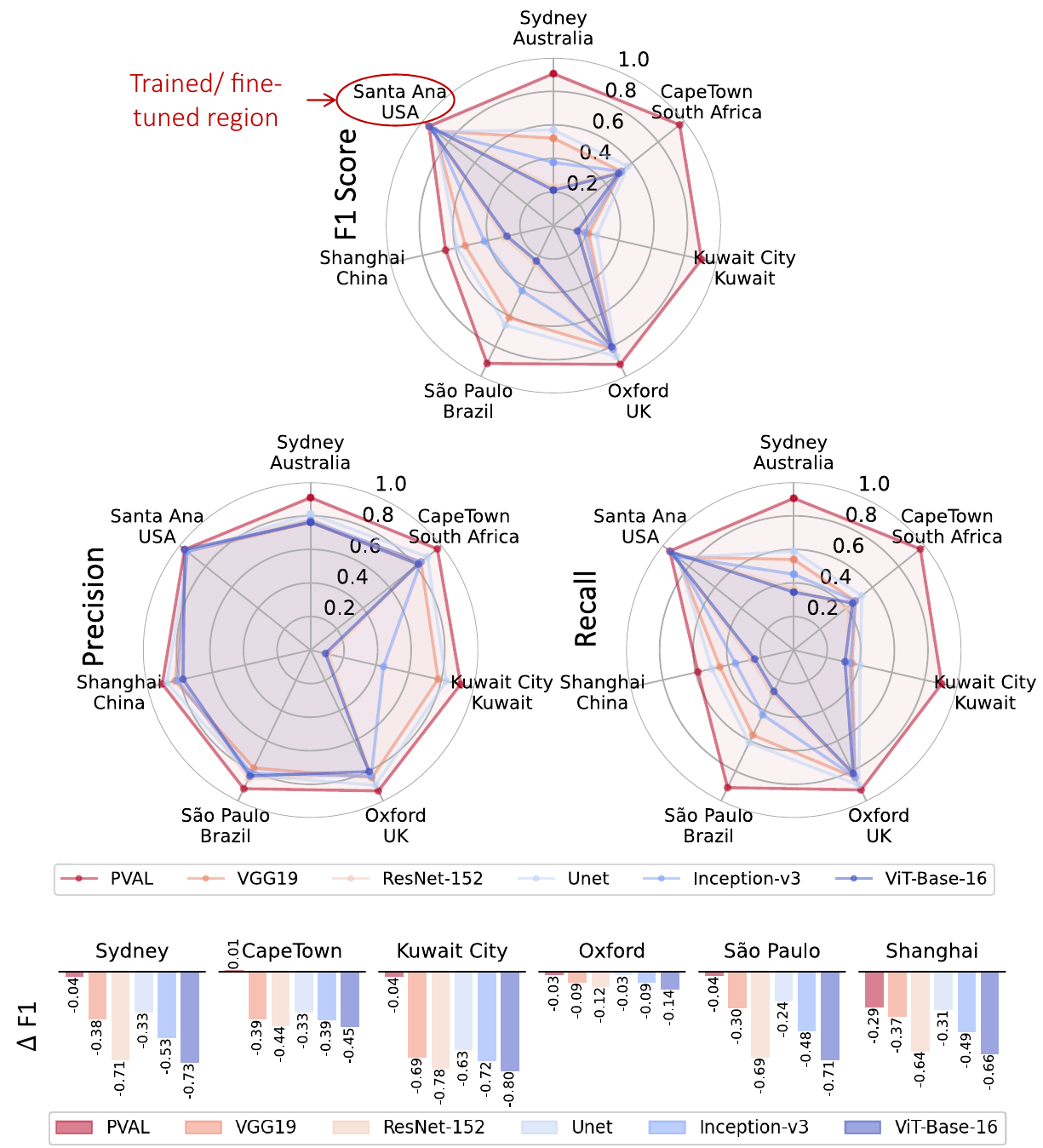}
\caption{
Top: Radar plots comparing model performance across seven regions for Precision, Recall, and F1-Score. The Santa Ana is the fine-tuning region. Bottom: $\Delta$F1 of each model across six unseen regions, where positive values denote improved transferability and negative values indicate reduced generalization.
}
\vspace{-1.5em}
\label{fig:Finetune_region}
\end{figure}

A core objective of this research is to assess the domain adaptation and generalization capabilities of the multimodal LLM for global-scale PV mapping. To this end, we conducted a comprehensive cross-domain evaluation, applying the PVAL exclusively on data from Santa Ana to six unseen regions.

The radar plots in Fig.~\ref{fig:Finetune_region} (top) highlight the model’s stability across domains. The PVAL’s performance (red line) maintains the broadest and most consistent region for F1-Score, Precision, and Recall, outperforming conventional vision models. In contrast, models such as U-Net and ResNet-152 rely on low-level visual features (e.g., ``dark rectangles on light-colored roofs”) that are strongly domain-dependent. When exposed to distinct rooftop materials: lighting conditions, or architectural designs, such as those in Kuwait City or São Paulo (see Fig.~\ref{fig:Framework}), these models exhibit sharp degradation due to their limited semantic understanding.

This performance decline is quantified in Fig.~\ref{fig:Finetune_region} (bottom) using the $\Delta$F1 metric. Although all models experience some drop when transferred to unseen regions, the PVAL’s decrease is consistently the smallest (e.g., -0.04 in Sydney, -0.03 in Oxford). This robustness stems from the LLM’s architecture and pretraining, which enables it to form a semantic understanding of photovoltaic panels independent of background textures or regional styles. Fine-tuning then adapts this generalized knowledge to the PV domain, yielding strong transferability with minimal loss under domain shift. 
\begin{figure}[htbp]
\centering
\includegraphics[width=1\columnwidth]{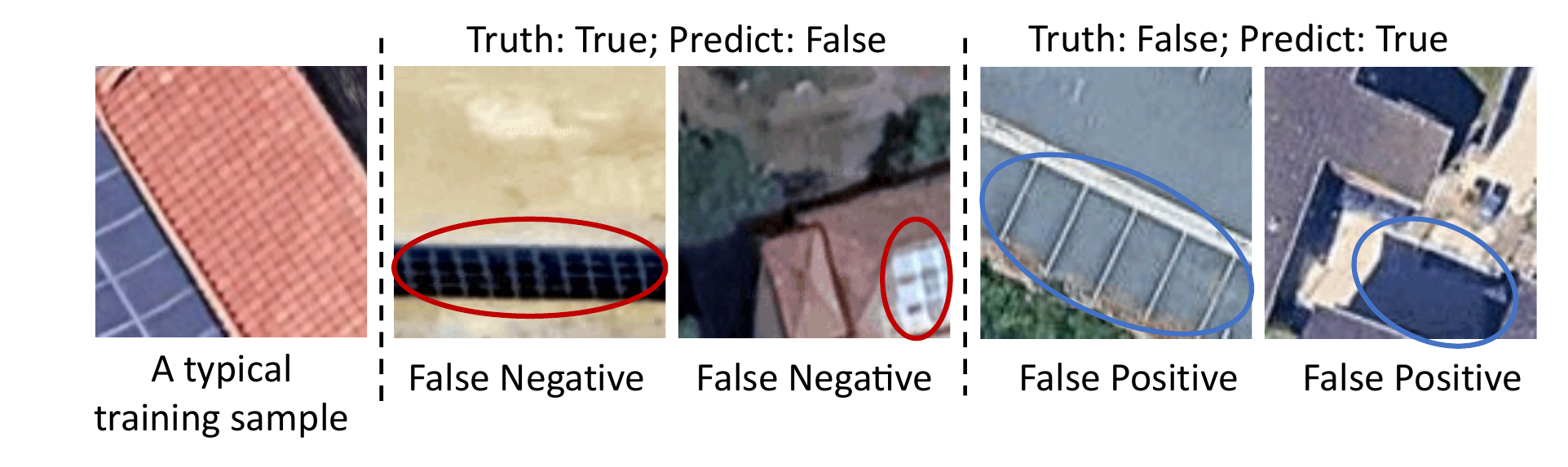}
\caption{
Example prediction errors of the ViT model. Middle: false negatives (missed PV panels); Right: false positives (non-solar structures misclassified as solar panels).
}
\vspace{-1em}
\label{fig:CaseStudy}
\end{figure}

Fig.~\ref{fig:CaseStudy} presents representative error cases from the ViT baseline. The false negatives demonstrate the model’s difficulty in identifying small or partially shaded PV panels, which differ from the clean, high-contrast panels seen during training. The false positives occur when roof features such as skylights or reflective surfaces mimic PV textures, leading to misclassification. These examples highlight the inherent limitations of conventional computer vision models that rely primarily on local texture cues, underscoring the advantage of multimodal LLMs that leverage semantic reasoning to disambiguate visually similar objects across varying domains.

\section{Conclusion}
This study explored and demonstrated that multimodal LLMs can achieve strong cross-domain generalization for global PV assessment. The PVAL framework, fine-tuned with structured prompts, maintains high accuracy across diverse regions and shows minimal $\Delta$\textit{F1} degradation under domain shift. By leveraging semantic reasoning rather than low-level visual cues, the LLM outperforms conventional vision models in robustness and adaptability. These findings highlight the potential of multimodal LLMs as scalable and transferable tools for global PV mapping and monitoring.

\bibliographystyle{IEEEtran}
\bibliography{IEEEabrv,reference}

\begin{thebibliography}{10}
\providecommand{\url}[1]{#1}
\csname url@samestyle\endcsname
\providecommand{\newblock}{\relax}
\providecommand{\bibinfo}[2]{#2}
\providecommand{\BIBentrySTDinterwordspacing}{\spaceskip=0pt\relax}
\providecommand{\BIBentryALTinterwordstretchfactor}{4}
\providecommand{\BIBentryALTinterwordspacing}{\spaceskip=\fontdimen2\font plus
\BIBentryALTinterwordstretchfactor\fontdimen3\font minus \fontdimen4\font\relax}
\providecommand{\BIBforeignlanguage}[2]{{%
\expandafter\ifx\csname l@#1\endcsname\relax
\typeout{** WARNING: IEEEtran.bst: No hyphenation pattern has been}%
\typeout{** loaded for the language `#1'. Using the pattern for}%
\typeout{** the default language instead.}%
\else
\language=\csname l@#1\endcsname
\fi
#2}}
\providecommand{\BIBdecl}{\relax}
\BIBdecl

\bibitem{guo2023graph}
M.~Guo, Q.~Cui, and Y.~Weng, ``Graph mining for classifying and localizing solar panels in distribution grids,'' in \emph{2023 Panda Forum on Power and Energy (PandaFPE)}.\hskip 1em plus 0.5em minus 0.4em\relax IEEE, 2023, pp. 1743--1747.

\bibitem{li2025external}
H.~Li, M.~Guo, M.~Ilic, Y.~Weng, and G.~Ruan, ``External data-enhanced meta-representation for adaptive probabilistic load forecasting,'' \emph{arXiv preprint arXiv:2506.23201}, 2025.

\bibitem{li2025exarnn}
H.~Li, M.~Guo, Y.~Weng, M.~Ilic, and G.~Ruan, ``Exarnn: An environment-driven adaptive rnn for learning non-stationary power dynamics,'' \emph{arXiv preprint arXiv:2505.17488}, 2025.

\bibitem{li2025latent}
H.~Li, C.~Xiao, M.~Guo, and Y.~Weng, ``Latent mixture of symmetries for sample-efficient dynamic learning,'' \emph{arXiv preprint arXiv:2510.03578}, 2025.

\bibitem{wu2022spatial}
J.~Wu, J.~Yuan, Y.~Weng, and R.~Ayyanar, ``Spatial-temporal deep learning for hosting capacity analysis in distribution grids,'' \emph{IEEE Transactions on Smart Grid}, vol.~14, no.~1, pp. 354--364, 2022.

\bibitem{yuan2022determining}
J.~Yuan, Y.~Weng, and C.-W. Tan, ``Determining maximum hosting capacity for pv systems in distribution grids,'' \emph{International Journal of Electrical Power \& Energy Systems}, vol. 135, p. 107342, 2022.

\bibitem{chen2023optimal}
M.~Chen, S.~Ma, Z.~Soltani, R.~Ayyanar, V.~Vittal, and M.~Khorsand, ``Optimal placement of pv smart inverters with volt-var control in electric distribution systems,'' \emph{IEEE Systems Journal}, vol.~17, no.~3, pp. 3436--3446, 2023.

\bibitem{simonyan2014very}
K.~Simonyan, ``Very deep convolutional networks for large-scale image recognition,'' \emph{arXiv preprint arXiv:1409.1556}, 2014.

\bibitem{bouazizintegrated}
C.~Bouaziz, M.~EL~Koundi, and G.~Ennine, ``Integrated solar panel detection and energy production estimation using unet and high-resolution imagery.''

\bibitem{luo2022solar}
S.~Luo, Y.~Weng, E.~Cook, R.~Trask, and E.~Blasch, ``Solar panel identification under limited labels,'' in \emph{2022 IEEE Power \& Energy Society General Meeting (PESGM)}.\hskip 1em plus 0.5em minus 0.4em\relax IEEE, 2022, pp. 1--5.

\bibitem{guo2023msq}
M.~Guo, M.~Guo, E.~T. Dougherty, and F.~Jin, ``Msq-biobert: ambiguity resolution to enhance biobert medical question-answering,'' in \emph{Proceedings of the ACM Web Conference 2023}, 2023, pp. 4020--4028.

\bibitem{guo2025solar}
M.~Guo and Y.~Weng, ``Solar photovoltaic assessment with large language model,'' \emph{Applied Energy}, vol. 402, p. 126835, 2025.

\bibitem{blasch2012high}
E.~Blasch, {\'E}.~Boss{\'e}, and D.~A. Lambert, \emph{High-level information fusion management and systems design}.\hskip 1em plus 0.5em minus 0.4em\relax Artech House, 2012.

\bibitem{josefsson2006base16}
S.~Josefsson, ``The base16, base32, and base64 data encodings,'' Tech. Rep., 2006.

\bibitem{he2016deep}
K.~He, X.~Zhang, S.~Ren, and J.~Sun, ``Deep residual learning for image recognition,'' in \emph{Proceedings of the IEEE conference on computer vision and pattern recognition}, 2016, pp. 770--778.

\bibitem{szegedy2016rethinking}
C.~Szegedy, V.~Vanhoucke, S.~Ioffe, J.~Shlens, and Z.~Wojna, ``Rethinking the inception architecture for computer vision,'' in \emph{Proceedings of the IEEE conference on computer vision and pattern recognition}, 2016, pp. 2818--2826.

\bibitem{dosovitskiy2020image}
A.~Dosovitskiy, ``An image is worth 16x16 words: Transformers for image recognition at scale,'' \emph{arXiv preprint arXiv:2010.11929}, 2020.

\end{thebibliography}
\end{document}